\documentclass{article}

\usepackage{microtype}
\usepackage{graphicx}
\usepackage{subfigure}
\usepackage{booktabs} 

\usepackage{hyperref}

\usepackage[accepted]{icml2023}

\usepackage{amsmath}
\usepackage{amssymb}
\usepackage{mathtools}
\usepackage{amsthm}

\usepackage[capitalize,noabbrev]{cleveref}

\theoremstyle{plain}

\theoremstyle{definition}

\theoremstyle{remark}

\usepackage{booktabs}       
\usepackage{amsfonts}       
\usepackage{nicefrac}       
\usepackage{microtype}      
\usepackage{xcolor}         
\usepackage{graphicx}
\usepackage{wrapfig}        
\usepackage{listings}
\usepackage{multirow}
\usepackage{makecell}

\def\Figref#1{Figure~\ref{#1}}
\def\Secref#1{Section~\ref{#1}}

\def\Tabref#1{Table~\ref{#1}}

\def\NAME{PyGlove}
\newcommand{\TT}[1]{\texttt{#1}}

\usepackage[textsize=tiny]{todonotes}

\setcitestyle{open={[},close={]},numbers}

\icmltitlerunning{PyGlove: Efficiently Exchanging ML Ideas as Code}

\begin{document}

\twocolumn[
\icmltitle{PyGlove: Efficiently Exchanging ML Ideas as Code}



\icmlsetsymbol{equal}{*}

\begin{icmlauthorlist}
\icmlauthor{Daiyi Peng}{brain}
\icmlauthor{Xuanyi Dong}{brain}
\icmlauthor{Esteban Real}{brain}
\icmlauthor{Yifeng Lu}{brain}
\icmlauthor{Quoc V. Le}{brain}
\end{icmlauthorlist}

\icmlaffiliation{brain}{Google Research, Brain Team}

\icmlcorrespondingauthor{Daiyi Peng}{daiyip@google.com}

\icmlkeywords{Machine Learning, ICML}

\vskip 0.3in
]



\printAffiliationsAndNoticeNoICML{}  

\begin{abstract}

The increasing complexity and scale of machine learning (ML) has led to the need for more efficient collaboration among multiple teams. For example, when a research team invents a new architecture like ``ResNet,'' it is desirable for multiple engineering teams to adopt it. However, the effort required for each team to study and understand the invention does not scale well with the number of teams or inventions. In this paper, we present an extension of our PyGlove library to easily and scalably share ML ideas. PyGlove represents ideas as symbolic rule-based patches, enabling researchers to write down the rules \emph{for models they have not seen}. For example, an inventor can write rules that will ``add skip-connections.''
This permits a \emph{network effect} among teams: at once, any team can issue patches to all other teams. Such a network effect allows users to quickly surmount the cost of adopting PyGlove by writing less code quicker, providing a benefit that scales with time. We describe the new paradigm of organizing ML through symbolic patches and compare it to existing approaches. We also perform a case study of a large codebase where PyGlove led to an 80\% reduction in the number of lines of code.

\end{abstract}

\section{Introduction}\label{sec:introduction}

Machine learning (ML) collaborations have increased in size dramatically over the last few years, making it more difficult to efficiently exchange code. Universities, GitHub projects, and especially technology companies can connect multiple researchers and engineers. Technology companies, in particular, are  often divided into several independent teams sharing a common codebase. As these teams produce discoveries, other teams need to incorporate them into their own code. Yet the specialization of teams and codebases can pose difficulties. The more common solution is to have each team scout for discoveries by other teams, then incorporate the discoveries into their own ML system. This can be time-consuming when inventions require specialized knowledge, are complex, or simply when there are too many of them (more than 5000 papers have been submitted  in 2022 to the ICML conference alone, for example). The alternative approach, where the inventors themselves apply their discovery in other's codebases presents additional challenges such as lack of access or sufficient documentation. Most importantly, these costs are incurred every time an exchange between two teams happens. The same invention ends up reimplemented multiple times, leading to poor scalability with the number of inventions. In this paper, we present {\NAME}, an extension of our previous work \cite{peng2020pyglove} that aims to simplify the scalable exchange of ideas as code.

\begin{figure}
  \begin{center}
    \includegraphics[width=\linewidth]{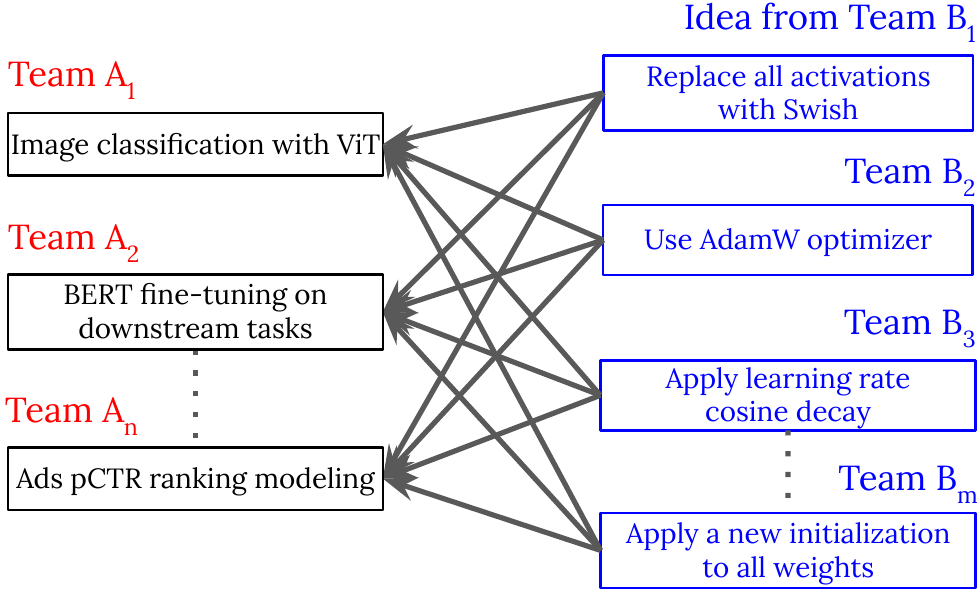}
  \end{center}
  \label{fig:network-effects}
\caption{
Applying $m$ new ML innovations to $n$ teams leads to $n{\times}m$ code changes.
{\NAME} expresses an ML innovation as a sharable rule that can be applied at once by each of the $m$ teams.
%
%
}
\end{figure}

\begin{table*}[!tb]
\centering
\setlength{\tabcolsep}{3.8pt}
\caption{
Statistics on the usage of PyGlove at Google demonstrate the positive impact of patching on code reuse: the number of experiments increases linearly, while the number of patching rules only grows slightly.
}
\vspace{2mm}
\begin{tabular}{c l c c c}
\toprule
        & Aspects & 2020   & 2021  & 2022 \\
\midrule
\multirow{5}{*}{\makecell{An internal \\ PyGlove-based \\ ML codebase}}
 & Accumulated \# of patching rules implemented                       & 70+      & 80+      & 80+      \\
 & Accumulated \# of experiments created with patching                & 10K+    & 18K+    & 25K+    \\
 & Accumulated \# of ML engineers on this codebase                      & 9        & 12       & 22       \\
 & Average \# of patching rules per experiment                    & 4.2      & 3.1      & 4.7      \\
 & Reduced lines of code per experiment                         & $\approx$80\% & $\approx$80\% & $\approx$80\% \\
\midrule
\multirow{2}{*}{\makecell{All PyGlove-based \\ projects}}
  & Total {\NAME}-based projects                 & 20+    & 40+   & 70+       \\
  & Total internal {\NAME} users                 & 100+   & 200+  & 1200+      \\ 
\bottomrule
\end{tabular}
\label{table:synthetic-data}
\end{table*}

{\NAME} allows a new invention to be applied in multiple places with minimal implementation work. The inventors themselves can edit other teams' code by broadcasting their discovery programmatically. At a high level, PyGlove works through annotations and rule-based patches. To make a codebase PyGlove-compatible, it must first be sprinkled with specially formatted, light-weight Python annotations that describe the code at an intuitive level. The annotations are the common language in which code-sharing will take place. Once that is done, code can be exchanged through rule-based patches that describe how the code being ported must be inserted elsewhere.
For example, consider a situation in which ``team A'' maintains an image classifier and ``team B'' independently comes up with a new convolutional layer that should improve most image classifiers. In the {\NAME} approach, team A can annotate their (pre-existing) code with statements akin to ``this is a convolution'', ``this is a nonlinearity'', and so on, and team B would annotate their new layer with statements akin to ``these are hyperparameters'' (see \Secref{sec:symbolic-patching}).
Upon team A learning about the new layer, A can write a one-line rule equivalent to ``replace all my convolutions with team B's layer'', as shown in \Figref{fig:symbolic_exp_sim_v1}.
Additionally, {\NAME} also allows a novel twist: team B can write the replacement rule themselves---such a rule would be analogous to ``in any image classifier, replace all convolutions with our layer''. Then any team with a {\NAME}-annotated image classifier can apply the rule produced by team B. This twist opens up future collaboration options through repositories of ML inventions, which we discuss in \Secref{subsec:patch-combinatorial-nature}.

\begin{figure*}[!ht]
    \centering
    \vspace{2mm}
    \includegraphics[width=\textwidth]{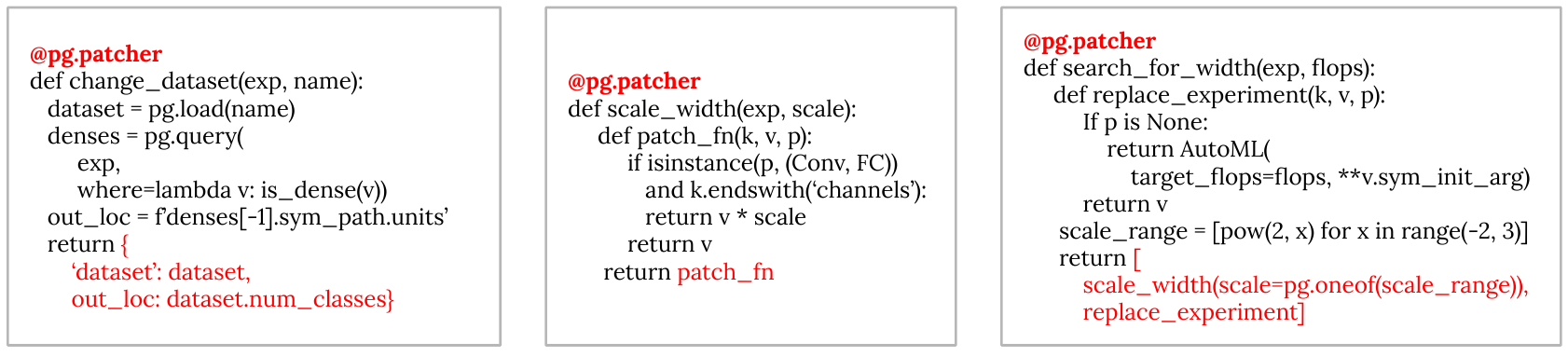}
    \vspace{-4mm}
    \caption{
Three more complex ideas are expressed via {\NAME} patching rules.
(Left) To change the dataset for a classification task, we need to change both the dataset and the model header's width to align with the number of classes.
This is done by querying the location of the last Dense layer, and returning the new value for both the dataset and the dense units associated with their locations as an update to the experiment.
(Middle) To multiply a scaling factor on the channels of the Conv and FC layer, the patching rule is defined by a function for transforming each node in the object tree, which takes the location, value and parent of current node as input, and modifies the `channels' member of Conv and FC classes. (Right) A upgraded version of previous example, which uses AutoML for scaling the model with a target FLOPs. To do so, it generates a search space of scale factors, and connects the search space to an AutoML trainer, which will automatically figure out the best choices.
}
    \label{fig:symbolic_exp_complex_v1}
\end{figure*}

We highlight that the convolution-layer-exchange example is a particularly simplistic one chosen for illustration purposes. More complex examples can be found in \Figref{fig:symbolic_exp_complex_v1}. Further, PyGlove's rule-based approach is not constrained to the sharing of architectural changes but applies to all aspects of the ML pipeline, such as data augmentation, training algorithm, or meta-learning. In particular, as ML hardware improves, there is a recurring need to scale up model capacity. Empirical and theoretical rules have been invented to address this problem~\cite{sussman2007building,tan2019efficientnet,yang2021tuning}. Such rules could be broadcasted throughout an organization or the community, saving valuable engineer time.

PyGlove's adoption cost can be quickly offset by its benefit due to a ``network effect'' among teams (\Figref{fig:network-effects}). The adoption cost is the (fixed) effort of annotating a codebase, in which little coding is needed beyond the new annotations themselves. As these are standard Python annotations, the bulk of the original code remains untouched. The benefit of PyGlove, on the other hand, is reaped every time an idea is exchanged by teams, as follows. 
Without PyGlove, $m$ innovations applied to $n$ team projects result in $m \times n$ work; with PyGlove, each innovation requires writing a PyGlove rule ($m$ rules) and each team project is responsible for PyGlove annotations in their own model ($n$ models), resulting in only $m + n$ work, as the rule application is trivial. 
In all these cases, our rule-based approach stands in contrast to current methods, which typically require multiple in-place edits that do not scale well with the size of the model or with the number of practitioners in the community.
These comparisons are detailed in \Secref{sec:pyglove-analysis} and quantitative real-world examples can be found in \Tabref{table:synthetic-data}. In particular, our case study of one large codebase showed an 80\% reduction in the number of lines of code due to the adoption of PyGlove.\looseness=-1

The reason why {\NAME} can be applied broadly to all aspects of ML code---and indeed beyond ML---is its principled symbolic programming design~\cite{mccarthy1960recursive}.
Under this paradigm, {\NAME}-annotated Python objects become editable symbols, and {\NAME} rules are meta-programs that act on these symbols.
The rules are also symbols themselves, allowing compositionality similar to that of a LISP program's ability to self-edit~\cite{steele1990common}.
In \Secref{sec:symbolic-program-ml}, we describe symbolic programming and explain the {\NAME} design.\looseness=-1

To summarize, we present:

\begin{itemize}
    \item A method for efficiently and scalably sharing complex ML ideas as code using symbolic patches;
    \item An illustration of how symbolic programming can be used throughout the ML development process;
    \item The open-sourced {\NAME} library\footnote{\url{https://github.com/google/pyglove}} and supplementary code used in this paper\footnote{\url{https://github.com/google/pyglove/blob/main/docs/notebooks/ml/efficiently_exchange_ml_ideas_as_code.ipynb}}.
\end{itemize}





\section{Expressing ML Ideas Through Patching }\label{sec:symbolic-patching}

Machine learning (ML) engineering often consists in the iterative improvement of an ML \emph{setup}. In this paper we define a \emph{setup} as the collection of components that produce a usable ML model, including the training data, its preprocessing, the model itself, the training algorithm, and the validation process. Whether for production or experimentation purposes, we can think of each improvement step as the editing of an initial setup by applying conceptual rules representing ideas, to produce a final setup. For example, simplistically speaking, if an ML practitioner wants to replace all convolution layers with separable convolution layers in a neural network, they would have to locate and replace each convolution unit in the network’s source code manually. This manual process requires an understanding of the different components and their interactions. 

\begin{figure}[!ht]
    \centering
    \vspace{-1mm}
    \includegraphics[width=\linewidth]{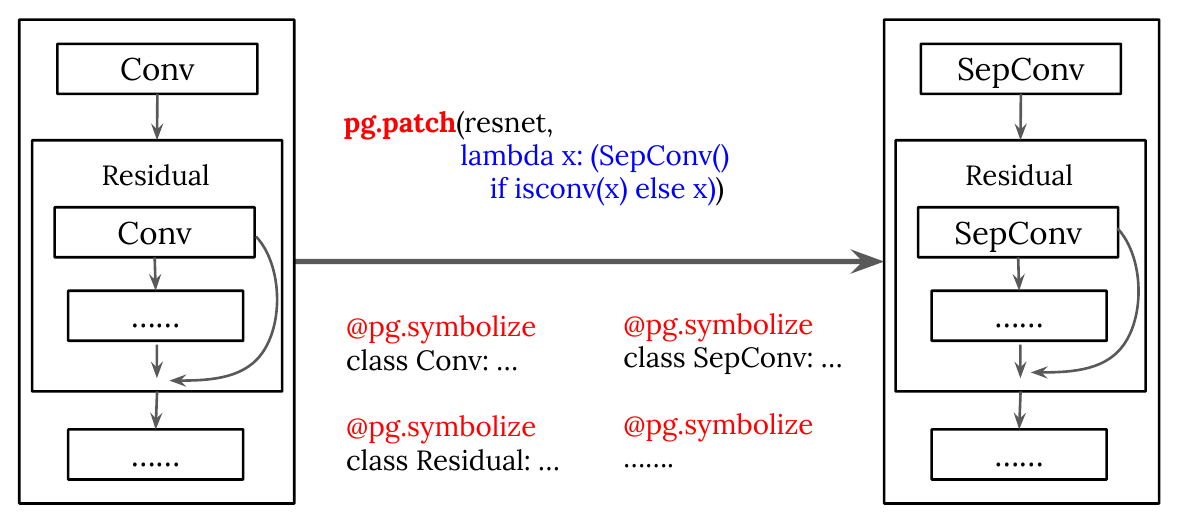}
    \vspace{-1mm}
    \caption{
Team B's new invention \TT{SepConv} is developed agnostic to Team A's code, but its rule \TT{apply\_sepconv} can be patched to upgrade Team A's experiment into \TT{new\_resnet}.
In this code, \TT{pg.symbolize} is the annotation that adds the editable trait to regular Python classes, and \TT{pg.patch} is the operation that applies a rule to an object of an annotated class.
}
    \label{fig:symbolic_exp_sim_v1}
    \vspace{-1mm}
\end{figure}


To automate this process, we need the ability to programmatically manipulate ML programs. This not only allows for the implementation of an idea but also enables the application of the same idea to other ML setups \emph{without modifying their source code}.
To achieve this, we developed {\NAME}, which allows for the expression of rules embodied in the manual process and encapsulates them into reusable units across different ML programs. This enables not only the implementation of complex ideas through direct manipulation of ML programs but also the capture of laborious daily ML practices through programmatic expression of rules, or \emph{patches}. For example, the aforementioned task of implementing team B's latest convolution invention in team A's setup can be achieved programmatically by applying a single-line patch, as illustrated in \Figref{fig:symbolic_exp_sim_v1}.

\subsection{The Expressivity of Patching}   \label{subsec:patch-power}

The ease of expressing rules using symbolic patches is not limited to simple ideas, but also applies to more complex concepts. {\NAME} allows for direct manipulation ~\cite{hutchins1985direct} of program parts in an ML setup, whether by location or by rule. For instance, as shown in ~\Figref{fig:symbolic_exp_complex_v1}, {\NAME} can be used to change datasets, adjust the width of neural models, and even automate model scaling through AutoML.

Symbolic patches, are beneficial not only for their ease of use in expressing ideas but also for their ability to be shared and easily applied to different ML setups. {\NAME} facilitates the sharing of patches through a URI-like string, which is human-readable and can be uniquely identified with a global name. For instance, a patch for an ML setup to use ImageNet~\cite{deng2009imagenet} dataset can be represented as \TT{change\_dataset?name=imagenet}.

\subsection{The Combinatorial Nature of Patching}   \label{subsec:patch-combinatorial-nature}

\begin{figure}[!thb]
  \vspace{1mm}
  \begin{center}
    \includegraphics[width=0.55\linewidth]{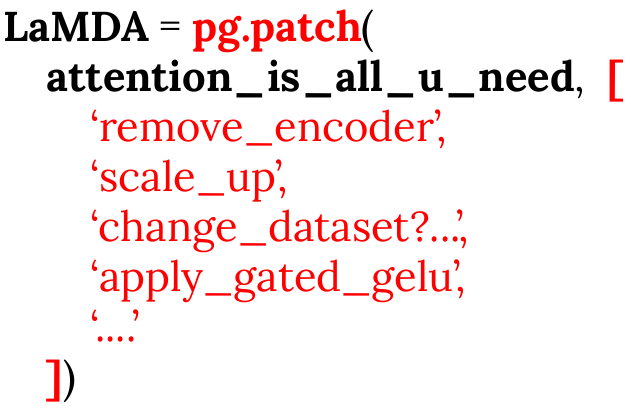}
  \end{center}
  \vspace{-4mm}
\caption{
The LaMDA paper~\cite{thoppilan2022lamda} can be obtained by applying a composition of patches on the Transformer paper~\cite{vaswani2017attention}.
}
\vspace{-4mm}
\label{fig:patch_composition}
\end{figure}



Furthermore, symbolic patches can be assembled to create more complex ideas. For instance, the recent large-scale NLP algorithm LaMDA~\cite{thoppilan2022lamda} can be obtained by applying a combination of common patches to the Transformer~\cite{vaswani2017attention} architecture, as we have illustrated in \Figref{fig:patch_composition} with the use of {\NAME}.

\begin{figure}[!h]
  \begin{center}
    \includegraphics[width=\linewidth]{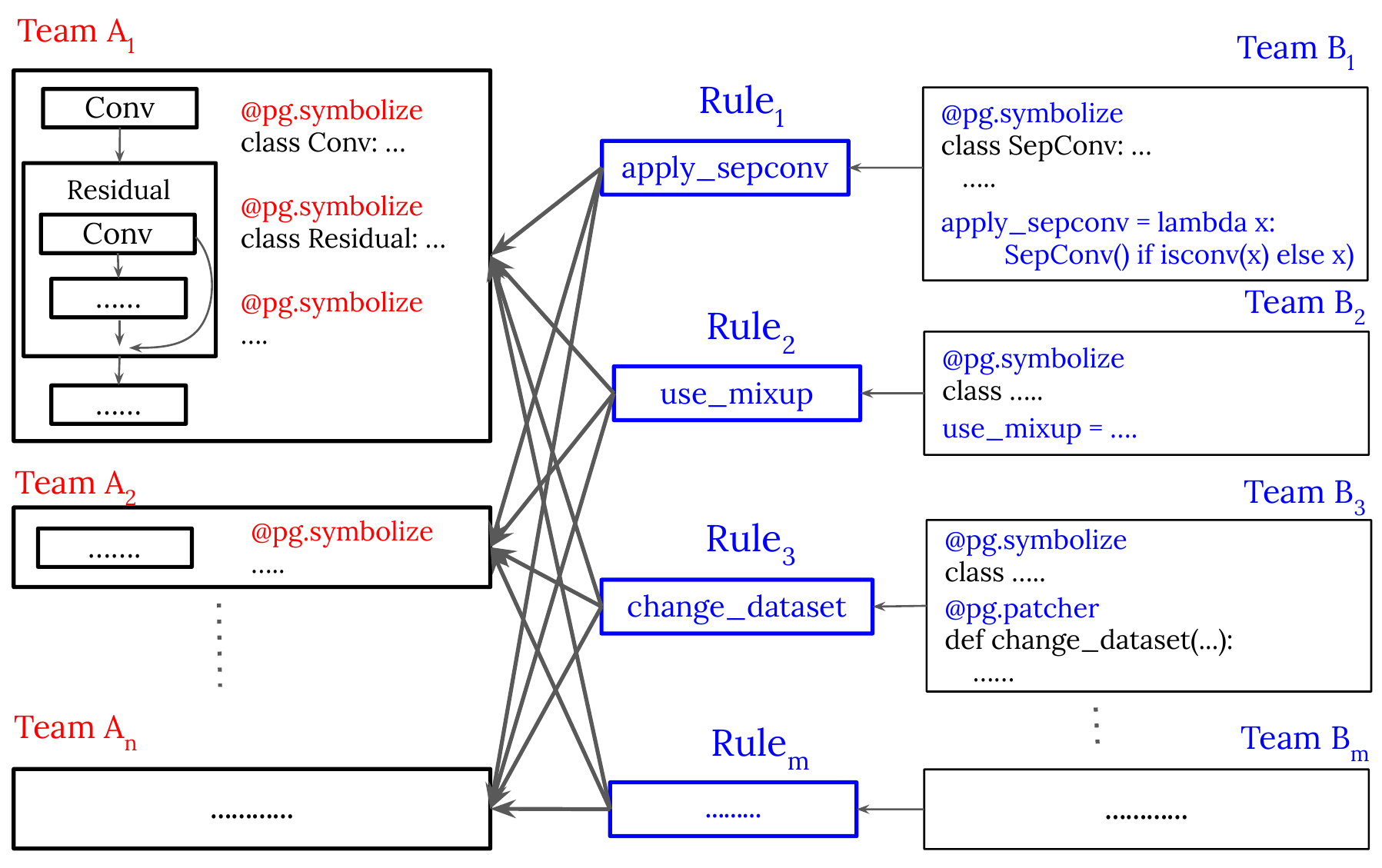}
  \end{center}
\caption{
Without {\NAME}, patching $m$ changes into $n$ teams requires $n\times{m}$ modifications to the code.
With {\NAME}, each $A_{i}$ team annotates their own code and each $B_{j}$ team writes their own rules.
As a result, patching all changes takes $n+m$ changes only.
}
\label{fig:deep_network_effect}
\end{figure}

The ability to compose patches is crucial for comprehensively exploring ML ideas. As shown in \Figref{fig:deep_network_effect}, research teams (Teams $B_{i}$) can share their latest discoveries with product teams (Teams $A_{i}$) by encapsulating them as patching rules using {\NAME}. Once a rule is created, it can be easily applied to any product team's codebase that is annotated in a compatible way. This means that a research team's idea can be adopted by many product teams without the need for manual integration each time, and product teams can quickly try out multiple research ideas. Additionally, the combination of multiple patches can further increase the potential for a network effect (see \Secref{sec:discussion}).



\subsection{Statistical Analysis for Patching's Effectiveness}

We have several non-public real-world projects at Google that have incorporated symbolic patching into their day-to-day ML practices, resulting in the conduct of tens of thousands of experiments over the last three years. The use of symbolic patching enables the scaling up of experiments without proportionally increasing development costs in large ML codebases, as demonstrated by the first project that adopted PyGlove which has over 50,000 lines of code.

\Tabref{table:synthetic-data} presents the usage statistics of the project, showing that by incorporating a slowly growing number of new patching rules, the total number of unique experiments launched by ML engineers on a daily basis has increased linearly. Furthermore, the development cost has been significantly reduced, and this impact has been consistently growing each year.





\section{Symbolic Programming for Machine Learning}\label{sec:symbolic-program-ml}

As we demonstrate the power and generality of symbolic patching for encapsulating ML ideas, its essential programming paradigm---symbolic programming---has a broader range of uses throughout the entire machine learning development process. Before delving deeper into its other applications, we will provide an overview of what symbolic programming is, its usefulness in machine learning, and the specific capabilities offered by {\NAME} in this area.


\subsection{Why Symbolic Programming Is Suitable for Machine Learning}

Symbolic programming, which originated from LISP \cite{steele1990common}, is a programming paradigm in which a computer program can manipulate its own components as if they were plain data~\cite{mccarthy1960recursive}.
Symbolic programming is useful in ML since ML is overloaded with a constant need for exploring new ideas based on existing ML setups, which requires frequent modification of ML programs.

Modifying programs can be done through editing the source code or by manipulating the program during runtime. However, using source code modification is not ideal for machine learning due to the need for flexibility in exploring new ideas and conducting various experiments. For instance, when users create new setups by copying and modifying an existing one, it becomes difficult to keep up with the changes if the original setup is updated.

In contrast, symbolic programming allows for the manipulation of a program during runtime without modifying the source code. This approach offers the ability to add new elements to existing ML setups in a dynamic manner without the need for managing source code changes. Additionally, it provides a direct link between ideas and implementation by allowing manipulation of pre-existing concepts.



\subsection{PyGlove's Symbolic Programming Solution}

\begin{figure}[!t]
\vspace{2mm}
  \begin{center}
    \includegraphics[width=\linewidth]{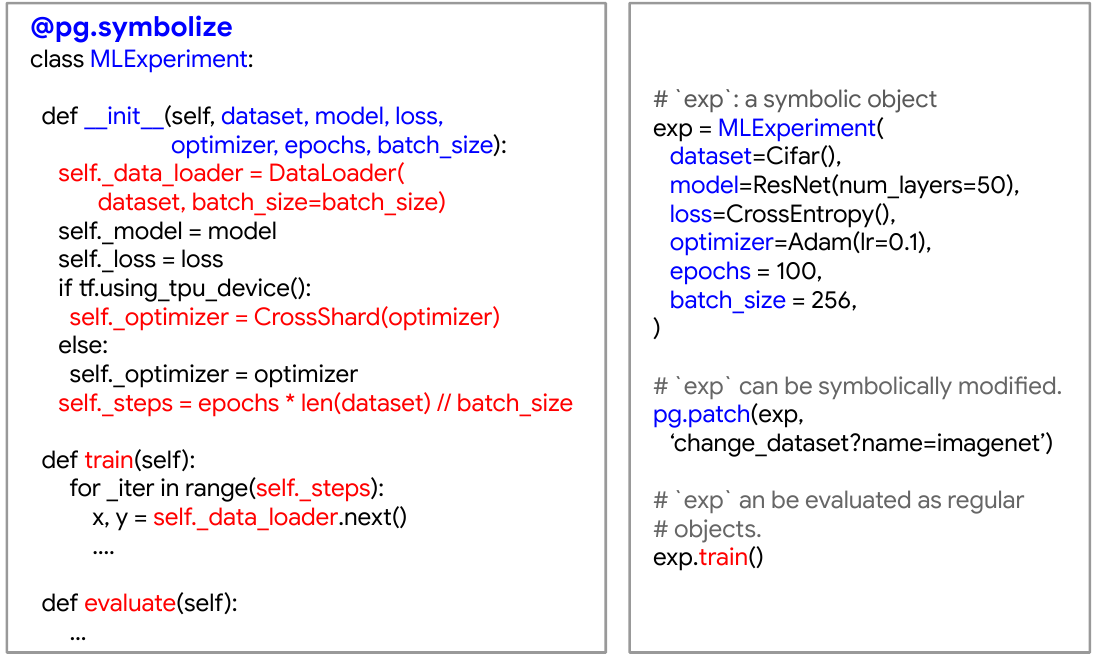}
  \end{center}
  \vspace{-2mm}
\caption{
Creating a symbolic class with a single-line of code (Left) and operating on its object (Right) 
}
\vspace{-3mm}
\label{fig:symbolic_class_object_def}
\end{figure}

PyGlove's symbolic programming solution involves the use of \emph{symbolic objects}, which are both symbols and objects. A symbolic object can be manipulated as a symbol while also being evaluated as a regular object. A symbolic object is an instance of a symbolic class, which can be created using the \TT{pg.symbolize} annotation. \Figref{fig:symbolic_class_object_def} illustrates the definition of a symbolic class \TT{MLExperiment} and its object \TT{exp}. 

\begin{figure}[!t]
  \begin{center}
    \includegraphics[width=0.80\linewidth]{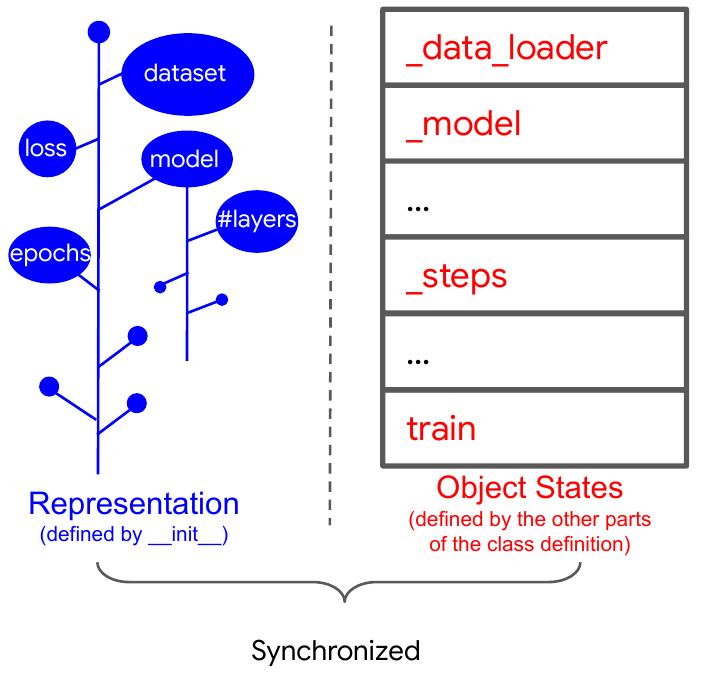}
  \end{center}
\caption{
A symbolic object's representation is defined by its constructor signature, while its object states are decided by the logic of execution. When the representation of a symbolic object changes, its object states will be reset, hence they are always consistent.
}
\vspace{-4mm}
\label{fig:symbolic_object_repr}
\end{figure}

\begin{figure}[!ht]
    \centering
    \includegraphics[width=0.89\linewidth]{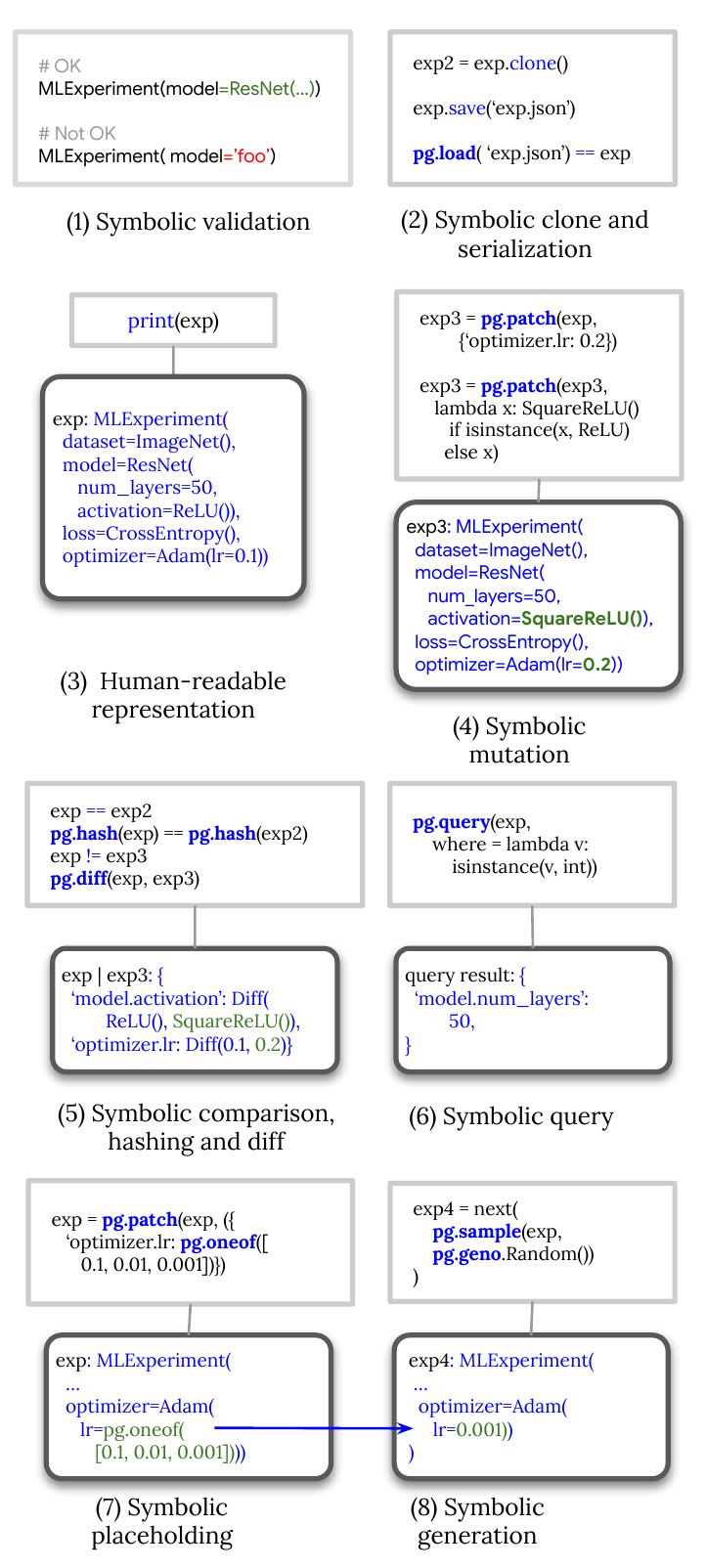}
    \caption{
    Common symbolic operations. A symbolic object can be validated, inspected, compared, cloned, serialized, mutated, place-held and generated with ease.
    }
    \label{fig:symbolic_ops}
\end{figure}

A symbolic object has several key characteristics, including the preservation of its construction information (type, constructor arguments) which defines its symbolic representation. This representation is identical between its source and runtime forms, allowing for the direct manipulation~\cite{hutchins1985direct} of concepts through their symbolic representations. Additionally, a symbolic object is also a program entity that exists at runtime, with data members and methods, behaving like a regular object (\Figref{fig:symbolic_object_repr}). The combination of being symbolic and being an object allows a symbolic object to benefit from both object-oriented programming and symbolic programming. While the former enables modular and extensible software design, the latter provides the power of meta-programming through manipulations. \Figref{fig:symbolic_ops} illustrates the operations that can be applied to symbolic objects.

\subsection{How Symbolic Objects Serve Machine Learning}



\begin{figure*}[t!]
    \centering
    \includegraphics[width=\textwidth]{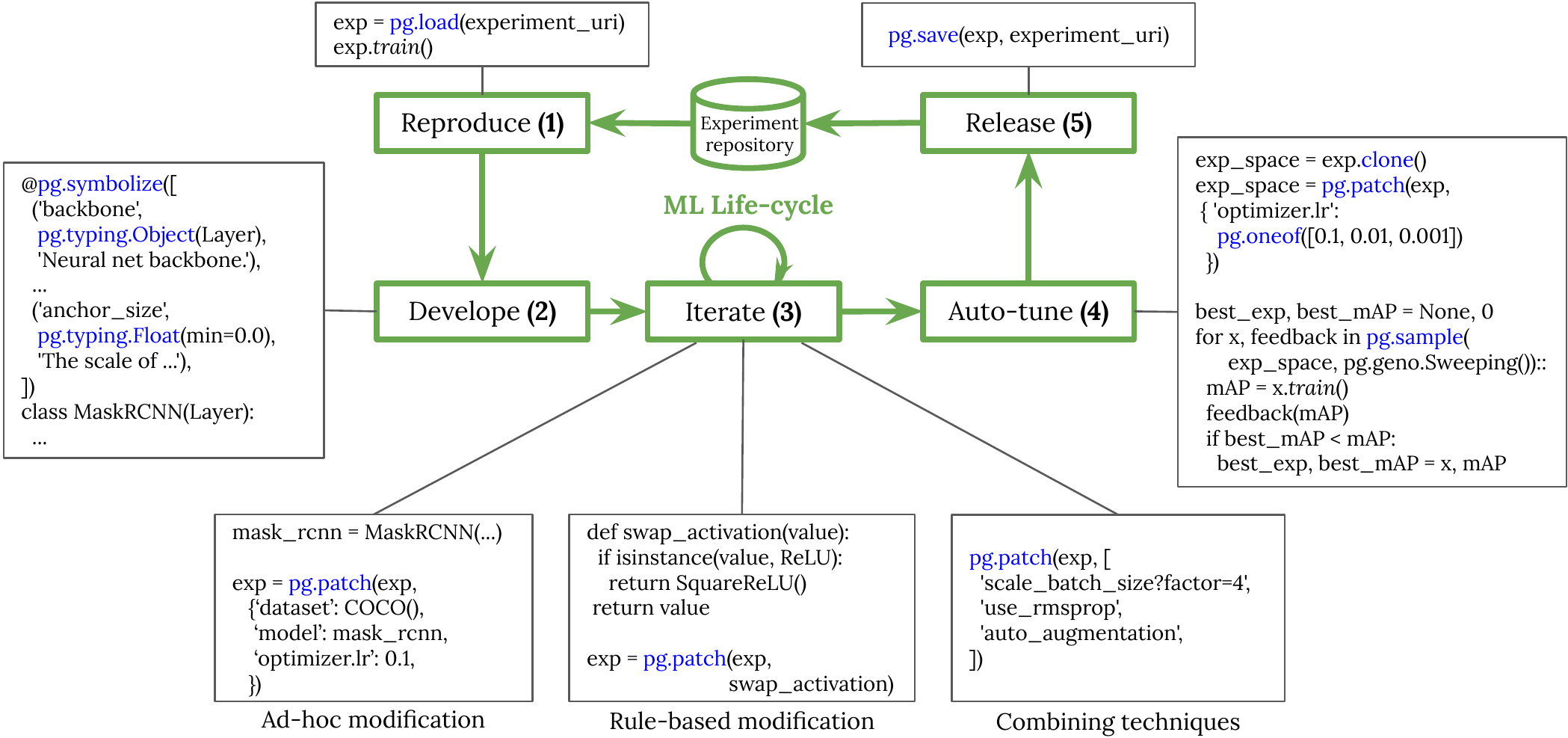}
    
    \vspace{2mm}
    \caption{
    {\NAME} serves the entire ML life-cycle.
    {\NAME} simplifies the engineering efforts involved from the birth of a research idea to its deployment  while making the whole or parts of an experiment reusable.\looseness=-1
    }
    \label{fig:ml_cycle}
    \vspace{2mm}
\end{figure*}

By examining the role of symbolic objects in machine learning, we can see the benefits they offer. Firstly, the ability to easily change the values of symbolic objects eliminates the need for separate hyperparameter containers, making the development and utilization of ML components more streamlined and consistent. Additionally, the use of symbolic operations allows for efficient experimentation in machine learning using minimal code. \Figref{fig:ml_cycle} demonstrates the symbolic operations used throughout the machine learning development process, based on the \TT{MLExperiment} class shown in \Figref{fig:symbolic_ops}.

The ability to reproduce an experiment (\Figref{fig:ml_cycle}(1)) is facilitated by obtaining the experiment object and evaluating it (e.g. by calling its \TT{train} method). As symbolic objects are serializable, ML setups can be loaded from a file or an experiment repository. As previously discussed in \Secref{sec:symbolic-patching}, iterating an experiment (\Figref{fig:ml_cycle}(3)) involves updating its setup with existing or new components. When new components are needed, they can be created in the same way as regular components, with the added step of applying the \TT{pg.symbolize} decorator (\Figref{fig:ml_cycle}(2)). While it is possible to manually iterate experiments, it can be time-consuming to do so, especially when a large number of modifications are being made. Symbolic placeholding simplifies automatic tuning (\Figref{fig:ml_cycle}(4)) by allowing the replacement of values to be tuned with search space specifications. This enables the generation of specific experiments from a larger space, controlled by tuning algorithms. This process works for both hyperparameters and neural architectures, as well as other experiment components such as losses, optimizers, data augmentation policies, etc. Finally, when an experiment is ready for deployment (\Figref{fig:ml_cycle}(5)), it can be serialized into a file or saved in an experiment repository. Not only the entire experiment, but also its individual parts, can be saved. Additionally, new components and patching rules can be added as standard ML modules.


\section{An Analysis of PyGlove's Approach}\label{sec:pyglove-analysis}

The core of {\NAME}'s approach to ML is encapsulating the engineering work required to move from one experiment to the next. This approach is based on the understanding that brand new ML building blocks are scarce, but their combinations and interactions are plentiful. {\NAME}'s solution allows for easy expression, reuse, and sharing of ideas through its flexible interfaces. Additionally, {\NAME} expands upon the object-oriented paradigm by incorporating symbolic programmability, enabling manipulation of ML experiment objects at runtime through rules, thus encapsulating the entire machine learning process rather than just its output.



\subsection{Comparison to Existing Solutions}

\begin{figure}[!hbt]
  \vspace{-3mm}
  \begin{center}
    \includegraphics[width=0.9\linewidth]{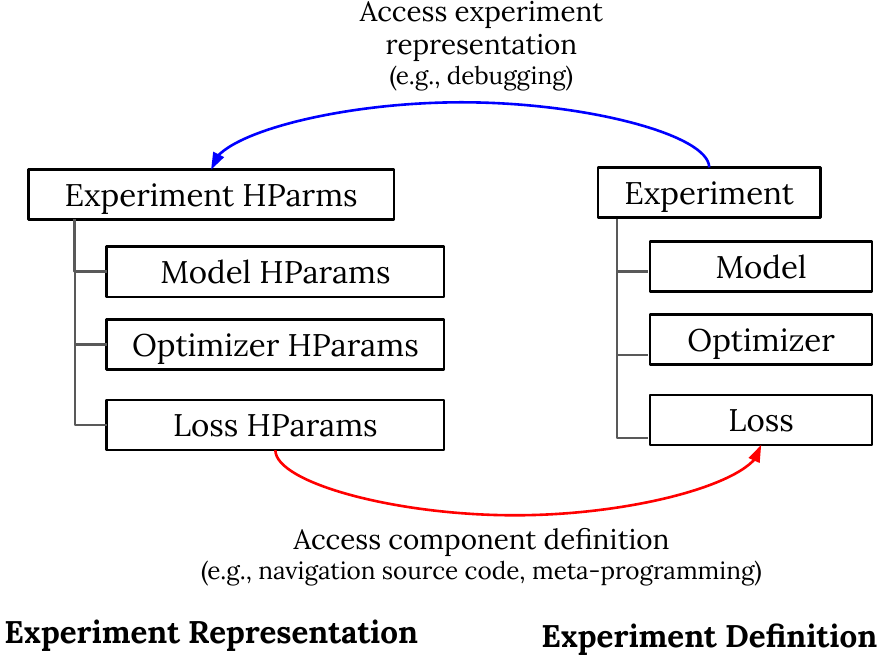}
  \end{center}
\caption{
The separation of experiment representation (Left) and experiment definition (Right) in a non-symbolic ML program. Experiment representations are usually written in configurations, while experiment definitions are expressed as source code of ML components. Builder pattern~\cite{gamma1993design} is usually involved in creating the ML component instances out of the experiments representations.
}
\label{fig:hp_container}
\end{figure}

The nature of ML, which centers around representations, has led to the development of solutions that allow for configuring ML programs using data-like entities, enabling modifications during experimentation without altering the source code. Current solutions~\cite{shen2019lingvo, tensorflowmodelgarden2020, wu2019detectron2} primarily utilize a configuration system, where ML components and their configurations are distinct entities. PyGlove, on the other hand, treats ML experiments as mutable symbolic objects, which combines representation and definition, illustrated in \Figref{fig:hp_container}.\looseness=-1



It is common for modifications to be made on ML setups by humans rather than rules based on configurations. This is because rule-based transformations often require access to runtime information, which is difficult to obtain when the experiment representation and program definition are separate. For example, when trying to change a dataset (\Figref{fig:symbolic_exp_complex_v1}-1), it may be difficult to retrieve the necessary information (e.g. \TT{num\_classes}) from the configuration alone. Additionally, rules are cost effective only when it can be widely used. The separation of representation and definition can lead to different schemas for the same concept, making it difficult for rules to be applied widely. However, in PyGlove's case, the program definition itself serves as a representation, leading to more uniform experiment representations that can be easily handled by rules.


The separation of experiment representation and its program definition not only makes it hard to write and use rules, but also causes inconvenience during the development of ML. This is because it requires synchronization between the two from time to time. This separation creates three common issues:

\begin{itemize}
    \item Developers need to modify the representation of a component when its definition is updated, and vice versa. This includes, but is not limited to, documentation, validation rules, and logic for creating sub-components.
    \item When working on an experiment representation, automation tools that need to access the component definitions cannot function properly, such as auto-completion and code navigation.
    \item When debugging a running experiment, it can be challenging to map errors back to the experiment representation, especially when dealing with complex compositions of the same component type. For example, identifying which Conv layer caused an input shape mismatch.
\end{itemize}

\subsection{Caveats and Limitations}

{\NAME} uses a mutable programming paradigm, which comes with its own set of caveats and limitations. It is widely acknowledged that mutable programs have three main drawbacks:
(1) They are harder to program safely, as objects can change, and extra care must be taken to ensure that program state remains consistent during mutations. (2) They are harder to reason about, as downstream code can update objects created earlier, making it more difficult to understand what is happening by simply looking at the code. (3) They are less performant, as mutable objects require additional instructions for bookkeeping the arguments in order to handle mutation, making the creation of symbolic objects slower than immutable objects. Additionally, using mutable objects for multi-threading is challenging due to the extensive locking required.
While these criticisms are generally valid, we argue that they do not have a significant impact in the case of ML.
ML languages provide a compilation step that separates the building of the setup and the training (e.g. the ``graph building'' in TensorFlow or the ``JIT'' step in JAX).
The ability to mutate the graph during the building phase provides a lot of power and flexibility, but once compiled, there is no need for further mutation. This has been supported by our use of {\NAME} in real-world research and product applications. 
Finally, the additional runtime cost incurred by symbolic object manipulations is negligible compared to the typical cost of ML training.

\section{Related Work}\label{sec:related-work}

PyGlove is a unique software library among the many existing ML libraries~\cite{abadi2016tensorflow,jax2018github,paszke2019pytorch, bergstra2010theano,jia2014caffe,chen2015mxnet,wu2019detectron2, tensorflowmodelgarden2020,shen2019lingvo}. Instead of adding specific features to ML programs, it introduces a new programming paradigm for ML by allowing experiments to be obtained from other experiments through the application of rules. This approach allows for easier expression of ideas in ML and enables reuse, sharing, and combination of rules to create a network effect among the ideas. PyGlove's approach to the ML engineering process is similar to the transform pipelines in compilers~\cite{bacon1994compiler, lattner2004llvm}, which take the intermediate representation (IR) of programs as input and apply transforms on top of it in chains. The IR in PyGlove is the symbolic representation of the object and the transforms are the patching rules. However, there are a few key differences between PyGlove and compilers: (1) In PyGlove, the rules are applied on the program itself, whereas in compilers the transforms are applied on other programs' IR. (2) The unity of IR and execution in PyGlove allows for modification and execution of a program to be combined, enabling more dynamic run-time behaviors.

PyGlove's solution is a form of symbolic programming~\cite{mccarthy1960recursive}, but it differs from other symbolic programming solutions, such as those found in LISP~\cite{steele1990common}, in that manipulations are done at the object level rather than the code definition level. This makes it more powerful as it allows the same code definition to have multiple instances that can be manipulated. Additionally, the manipulation is safer as it operates on the binding parameters of an object, rather than the lower-level code implementations, which are often self-sufficient. To be more specific, PyGlove's symbolic paradigm is a combination of symbolic programming and object-oriented programming (OOP) \cite{stefik1985object}. It allows a symbolic object to be manipulated as a symbol and evaluated as an object, thus taking advantage of the benefits of both programming paradigms. While the former allows manipulation of programs, the latter allows the development of reusable components through encapsulation and inheritance, as per the best practices of OOP.

In the field of ML, the most related work on using symbolic programming is \cite{peng2020pyglove}, which re-conceptualizes AutoML as a process of algorithm-based symbolic manipulations, and can be considered a specialized application of PyGlove. To expand its usefulness for more general ML tasks, this paper formalizes the symbolic object paradigm, highlights the importance of encapsulating rules, and emphasizes the network effect among them. Additionally, it expands the utilization of symbolic programming throughout the entire ML development process. As a result, this paper significantly broadens the scope of previous work.

\section{Discussion}
\label{sec:discussion}

Symbolic patching is a powerful tool for sharing machine learning ideas when all experiments use the same symbolic representations, which is typically the case within a shared codebase. In this paper, we have focused on this single-codebase scenario, which can be found within industry for example. Yet, this scenario does not apply to most academia or to the ML community as a whole, where different teams tend to build their own codebases fully indepedently. Scaling to the multi-codebase scenario poses some problems. In particular, participant teams would need to agree on a high-level interface. This requires community effort and usage of shared best practices in software design, and is beyond the scope of this paper. Nevertheless, we speculate that PyGlove offers new tools to help attain a shared high-level ML interface across codebases: (1) PyGlove works through code annotation rather than direct editing, which naturally permits building up the interfaces \emph{incrementally}. It is possible to annotate a codebase to only accommodate a particular rule patch of interest (e.g. one just published by another team). (2) PyGlove removes the need for the common paradigm of ``configuration objects'' (see \Secref{sec:pyglove-analysis}). Configuration objects have led to debate as to their format and conventions, preventing codebase convergence. Symbolic ML objects, on the other hand, are editable directly, eliminating the need for separate configuration. (3) The symbolic approach allows for compositionality, which in turns permits multi-level interfaces. For example, it is easy for the interface to simultaneously expose an image classifier as a whole, each layer within it, and the inputs and outputs within each layer. (4) The publication of useful patches can provide a strong incentive for codebase annotation, which may in turn encourage more patches, leading to positive reinforcement. We therefore hope that, in addition to sharing code within a codebase, PyGlove may also have a positive impact on collaboration across ML codebases in the future.

\section{Conclusion}

Machine learning is hindered by the inability to apply conceptual rules to different ML setups in a scalable manner. Even if the code and experiment definition of a paper are open-sourced, obtaining an ML setup from the paper and referential codebase is neither straightforward nor reusable, as it needs to be manually parsed and replicated on other experiments. In this paper, we have expanded {\NAME}~\cite{peng2020pyglove}'s symbolic capabilities from AutoML to ML, to address this problem. Our proposal encapsulates the process of evolving from ML setup A to setup B into a list of patching rules. These rules can be reused across ML setups, creating network effects among the experiments and rules. In addition to patching, we have also demonstrated how symbolic programming can serve the entire lifecycle of ML effectively. We have also compared {\NAME}'s solution with existing solutions. Through real-world research projects~\cite{dong2021autohas,zhou2022towards} that heavily rely on {\NAME}'s patching capability, we have seen the potential of how {\NAME} can change the way ML programs are developed, organized, and shared. We have open-sourced {\NAME} and look forward to it being extensively tested by the machine learning community.


\bibliography{abbreviation,ms}
\bibliographystyle{icml2023}



\end{document}


\maketitle

\begin{abstract}
X
\end{abstract}

\section{Explanation of Terminology}

\begin{table}[t!]
\centering
\setlength{\tabcolsep}{2.1pt}
\caption{
The glossary to explain the detailed meaning of some words.
}
    \begin{tabular}{p{0.3\linewidth} | l}
    \toprule
    Terminology              &  Explanations \\
    \midrule
    Machine learning program &  \\
    Program                  &  \\
    Representation           &  \\
    Meta-representation      &  \\
    \bottomrule
    \end{tabular}
\label{table:compare-setting}
\end{table}

\section{How SOOP Works?}

\bibliographystyle{plain}
\bibliography{abbreviation,reference}